\documentclass[letterpaper]{article} 
\usepackage{aaai25}  
\usepackage{times}  
\usepackage{helvet}  
\usepackage{courier}  
\usepackage[hyphens]{url}  
\usepackage{graphicx} 
\usepackage{natbib}  
\usepackage{caption} 
\usepackage{algorithm}
\usepackage{algorithmic}

\usepackage{newfloat}
\usepackage{listings}
\DeclareCaptionStyle{ruled}{labelfont=normalfont,labelsep=colon,strut=off} 
\floatstyle{ruled}
\newfloat{listing}{tb}{lst}{}
\floatname{listing}{Listing}
\usepackage{amsmath}
\usepackage{amsfonts,amssymb}
\usepackage{booktabs}
\usepackage{utfsym}
\usepackage{tabularx} 

\title{Attentive Eraser: Unleashing Diffusion Model's Object Removal Potential via Self-Attention Redirection Guidance}
\author{
       Wenhao Sun\textsuperscript{\rm 1}\equalcontrib, 
       Benlei Cui\textsuperscript{\rm 2}\equalcontrib, 
   Xue-Mei Dong\textsuperscript{\rm 1}\thanks{Corresponding author}, 
       Jingqun Tang\textsuperscript{\rm 3},
}

\affiliations{
       \textsuperscript{\rm 1}School of Statistics and Mathematics, Zhejiang Gongshang University, Hangzhou, China\\
    \textsuperscript{\rm 2}Alibaba Group,
Hangzhou, China\\
    \textsuperscript{\rm 3}ByteDance Inc.,
Hangzhou, China\\
    22020040141@pop.zjgsu.edu.cn,
    cuibenlei.cbl@alibaba-inc.com,
    dongxuemei@zjgsu.edu.cn,
    tangjingqun@bytedance.com,
    
}

\begin{document}
\maketitle

\begin{abstract}
Recently, diffusion models have emerged as promising newcomers in the field of generative models, shining brightly in image generation. However, when employed for object removal tasks, they still encounter issues such as generating random artifacts and the incapacity to repaint foreground object areas with appropriate content after removal. To tackle these problems, we propose \textit{Attentive Eraser}, a tuning-free method to empower pre-trained diffusion models for stable and effective object removal. Firstly, in light of the observation that the self-attention maps influence the structure and shape details of the generated images, we propose Attention Activation and Suppression (ASS), which re-engineers the self-attention mechanism within the pre-trained diffusion models based on the given mask, thereby prioritizing the background over the foreground object during the reverse generation process. Moreover, we introduce Self-Attention Redirection Guidance (SARG), which utilizes the self-attention redirected by ASS to guide the generation process, effectively removing foreground objects within the mask while simultaneously generating content that is both plausible and coherent. Experiments demonstrate the stability and effectiveness of Attentive Eraser in object removal across a variety of pre-trained diffusion models, outperforming even training-based methods. Furthermore, Attentive Eraser can be implemented in various diffusion model architectures and checkpoints, enabling excellent scalability. Code is available at \url{https://github.com/Anonym0u3/AttentiveEraser}.
\end{abstract}
\section{Introduction}
The widespread adoption of diffusion models (DMs) \cite{d1,s1,he2024mars,liu2024llm4gen} in recent years has enabled the generation of high-quality images that match the quality of real photos and provide a realistic visualization based on user specifications. This raises a natural question of whether the image-generating capabilities of these models can be harnessed to remove objects of interest from images. Such a task, termed object removal \cite{or1,lama}, represents a specialized form of image inpainting, and requires addressing two critical aspects. Firstly, the user-specified object (usually given as a binary mask) must be successfully and effectively removed from the image. Secondly, the mask area must be filled with content that is realistic, plausible, and appropriate to maintain overall coherence within the image.

Traditional approaches for object removal are the patch-based methods \cite{p1,g1}, which fill in the missing regions after removal by searching for well-matched replacement patches ($i.e.$ candidate patches) in the undamaged part of the image and copying them to the corresponding removal locations. However, such processing methods often lead to inconsistency and unnaturally between the removed region and its surroundings. In recent years, convolutional neural networks (CNNs) have demonstrated considerable potential for object removal tasks. However, CNNs-based methods \cite{s2,oleksii2019deep,lama} typically utilize a fixed-size convolutional kernel or network structure, which constrains the perceptual range of the model and the utilization of contextual information \cite{fang2023annotations,dbctnet,fang2025multi}. Consequently, the model's performance is sub-optimal when confronted with large-scale removal or complex scenes.

With the rapid development of generative models \cite{shenadvancing, zhang1, zhang2, zhang3, zhang4, wang2025pargo} in deep learning\cite{tang2022optimal,shen2023triplet,fang2024your,fang2025rethinking,liu2024pandora,liu2024envisioning,li2024frame}, a proliferation of generative models has been applied to object removal. Among these, the most common are generative adversarial network (GAN) \cite{gan}-based methods and DMs-based methods. GAN-based methods \cite{gb1,gb2} employ neural networks of varying granularity, with the context-focused module exhibiting robust performance and efficacy in image inpainting. However, their training is inherently slow and unstable, and they are susceptible to issues such as mode collapse or failure to converge \cite{ganbad1}. 

In current times, DMs have made new waves in the field of deep generative models, broken the long-held dominance of GANs, and achieved new state-of-the-art performance in many computer vision tasks \cite{shen2024imagdressing,shen2024boosting,shenadvancing,shen2024imagpose,zhao2024harmonizing}. The most prevalent open-source pre-trained model in DMs is Stable Diffusion (SD) \cite{sd}, which is a pre-trained latent diffusion model. To apply SD to the object removal task, fine-tuned from SD, SD-inpainting \cite{sd} was developed into an end-to-end model with a particular focus on inpainting, to incorporate a mask as an additional condition within the model. However, even after spending a considerable cost in terms of resources, its object removal ability is not stable, and it often fails to completely remove the object or generates random artifacts(as shown in Figure \ref{Qualitative}). An additional methodology entails guiding the model to perform object removal via prompt instruction \cite{gqa,Instructpix2pix}. The downside of this method is that to achieve a satisfactory result, these models often necessitate a considerable degree of prompt engineering and fail to allow for accurate interaction even with a mask. Additionally, they often necessitate substantial resources for fine-tuning.

\begin{figure}[t]
\centering
\includegraphics[width=0.9\columnwidth]{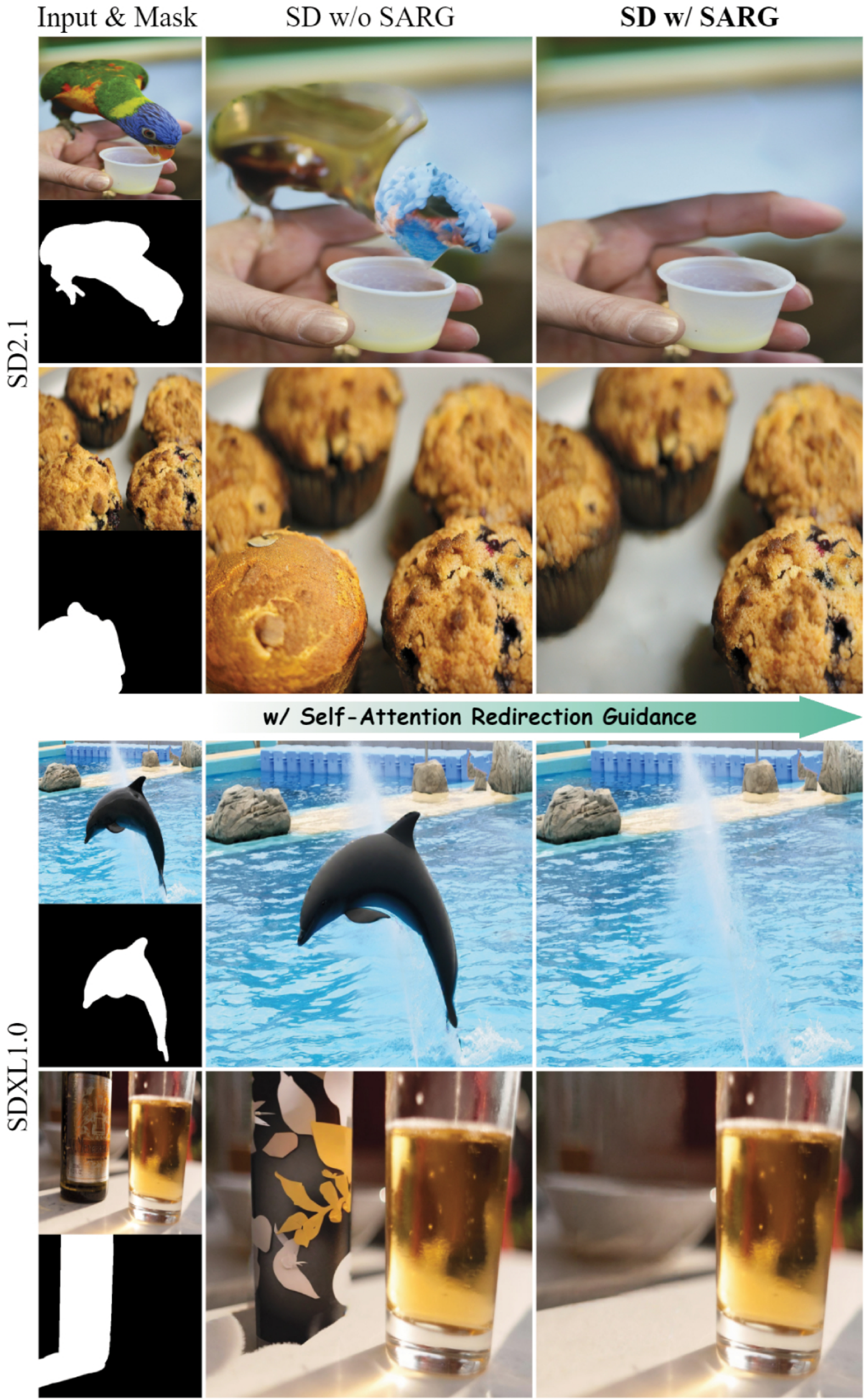} 
\caption{Qualitative comparison between Stable Diffusion (baseline) and self-attention redirection guided Stable Diffusion for object removal.}
\label{fig:1}
\end{figure}

To address these problems, we propose a tuning-free method, Attentive Eraser, a simple yet highly effective method for mask-guided object removal. This method ensures that during the reverse diffusion denoising process, the content generated within the mask tends to focus on the background rather than the foreground object itself. This is achieved by modifying the self-attention mechanism in the SD model and utilizing it to steer the sampling process. We show that when Attentive Eraser is combined with the prevailing diffusion-based inpainting pipelines \cite{diffedit,bld}, these pipelines enable stable and reliable object removal, fully exploiting the massive prior knowledge in the pre-trained SD model to unleash its potential for object removal (as shown in Figure \ref{fig:1}). The main contributions of our work are presented as follows:
\begin{itemize}
\item We propose a tuning-free method \textbf{Attentive Eraser} to unleash DM's object removal potential, which comprises two components: (1) \textbf{Attention Activation and Suppression (AAS)}, a self-attention-modified method that enables the generation of images with enhanced attention to the background while simultaneously reducing attention to the foreground object. (2) \textbf{Self-Attention Redirection Guidance (SARG)}, a novel sampling guidance method that utilizes the proposed AAS to steer sampling towards the object removal direction.
\item Experiments and user studies demonstrate the effectiveness, robustness, and scalability of our method, with both removal quality and stability surpassing SOTA methods.
\end{itemize}

\section{Related Works}
\subsection{Diffusion Models for Object Removal}
Existing diffusion model-based object removal methods can be classified into two categories, tuning-free \cite{zhao2024multi} vs. training-based \cite{fang2023you}, depending on whether they require fine-tuning or not. In the case of the training-based methods, DreamInpainter \cite{dreaminpainter} captures the identity of an object and removes it by introducing the discriminative token selection module. Powerpaint \cite{powerpaint} introduces learnable task prompts for object removal tasks. Inst-Inpaint \cite{gqa} constructs a dataset for object removal, and uses it to fine-tune the pre-trained diffusion model. There are other instruction-based methods achieving object removal via textual commands \cite{tb1,tb2,tb3}. In the case of the tuning-free methods, Blended Diffusion \cite{bld} and ZONE \cite{zone} perform local text-guided image manipulations by introducing text conditions to the diffusion sampling process. Magicremover \cite{magicremover} implements object removal by modifying cross-attention to direct diffusion model sampling. SuppressEOT \cite{liget} suppresses negative target generation by focusing on the manipulation of text embeddings. However, these methods can lead to artifacts in the final result or incomplete removal of the target due to the stochastic nature of the diffusion model itself and imprecise guiding operations. To address the above issues and to avoid consuming resources for training, we propose a tuning-free method SARG to gradually steer the diffusion process towards object removal.

\subsection{Sampling guidance for diffusion models}
Sampling guidance for diffusion models involves techniques that steer the sampling process toward desired outcomes. Classifier guidance \cite{G5} involves the incorporation of an additional trained classifier to generate samples of the desired category. Unlike the former, Classifier-free Guidance \cite{G6} does not rely on an external classifier but instead constructs an implicit classifier to guide the generation process. There are two methods that combine self-attention with guidance, SAG \cite{SAG} and PAG \cite{PAG}, which utilize or modify the self-attention mechanism to guide the sampling process, thereby enhancing the quality of the generated images. Our work is similar to PAG in that it modifies the self-attention map to guide sampling, but the purpose and approach to modification are different.

\section{Preliminaries}
\subsection{Diffusion Models}
DMs are a class of probabilistic generative models that learn a given data distribution $q\left(x\right)$ by progressively adding noise to the data to destroy its structure and then learning a corresponding inverse process of a fixed Markov chain of length T to denoise it. Specifically, given a set of data $x_0 \sim q\left(x_0\right)$, the forward process could be formulated by
\begin{equation}  
q\left(x_t \mid x_{t-1}\right)=\mathcal{N}\left(x_t ; \sqrt{1-\beta_t} x_{t-1}, \beta_t \mathbf{I}\right),
\end{equation}
where $t \in \{ 1, 2, \dots, T \}$ denotes the time step of diffusion process, $x_t$ is the noisy data at step $t$, $\beta_t \in [0,1]$ is the variance schedule at step $t$ and represents the level of noise. 

Starting from $x_T$, the reverse process aims to obtain a true sample by iterative sampling from $q\left(x_{t-1} \mid x_t\right)$. Unfortunately, this probability is intractable, therefore, a deep neural network with parameter $\theta$ is used to fit it:
\begin{equation}  
p_\theta\left(x_{t-1} \mid x_t\right) = \mathcal{N}\left(x_{t-1} ; \mu_\theta^{(t)}(x_t), \Sigma_\theta^{(t)}(x_t)\right),
\end{equation}  
With the parameterization
\begin{equation}  
\mu_\theta^{\left ( t \right ) }\left(x_t\right)=\frac{1}{\sqrt{\alpha_t}}\left(x_t-\frac{\beta_t}{\sqrt{1-\bar{\alpha}_t}} \epsilon_\theta^{\left ( t \right ) }\left(x_t\right)\right),
\end{equation}  
proposed by Ho\cite{d1}, a U-net \cite{unet} $\epsilon_\theta^{\left ( t \right ) }\left(x_t\right)$ is trained to predict the noise $\epsilon \sim \mathcal{N}(\mathbf{0}, \mathbf{I})$ that is introduced to $x_0$ to obtain $x_t$, by minimizing the following object:
\begin{equation}  
\min _\theta \mathbb{E}_{x_0, \epsilon \sim \mathcal{N}(\mathbf{0}, \mathbf{I}), t \sim \text { Uniform }(1, T)}\left\|\epsilon-\epsilon_\theta^{\left ( t \right ) }\left(x_t\right)\right\|_2^2,
\end{equation}  
After training, a sample $x_0$ can be generated following the reverse process from $x_T \sim \mathcal{N}(\mathbf{0}, \mathbf{I})$. 

\subsection{Self-Attention in Stable Diffusion}
Recent studies \cite{localizing,dreammatcher,Towards} have elucidated the significant role of the self-attention module within the stable diffusion U-net. It harnesses the power of attention mechanisms to aggregate features \cite{tang2022few,shen2023git,fang2023hierarchical}, allowing for a more nuanced control over the details of image generation.
Specifically, given any latent feature map $z\in \mathbb{R}^{h\times w \times c}$, where $h$, $w$ and $c$ are the height, width and channel dimensions of $z$ respectively, the according query matrix $Q_{self}\in \mathbb{R}^{\left ( h\times w \right ) \times d}$, key matrix $K_{self}\in \mathbb{R}^{\left ( h\times w \right ) \times d}$ and value matrix $V_{self}\in \mathbb{R}^{\left ( h\times w \right ) \times d}$ can be obtained through learned linear layers $\ell_{Q}$ , $\ell_{K}$ and $\ell_{V}$, respectively. The similarity matrix $S_{self}$, self-attention map $A_{self}$ and output $OP_{self}$ can be defined as follows:
\begin{equation}  
Q_{self} = \ell_{Q}\left ( z \right ), K_{self} = \ell_{K}\left ( z \right ), V_{self} = \ell_{V}\left ( z \right ),
\end{equation}  
\begin{equation}  
S_{self} = Q_{self}\left ( K_{self} \right )^{T} /\sqrt{d},
\end{equation}  
\begin{equation}  
A_{self} = \text{softmax}\left ( S_{self}\right ),
\end{equation}  
\begin{equation}  
OP_{self} = A_{self} V_{self},
\end{equation}
where $d$ is the dimension of query matrix $Q_{self}$, and the similarity matrix $S_{self}\in \mathbb{R}^{\left ( h\times w \right ) \times \left ( h\times w \right )}$ and self-attention map $A_{self}\in \mathbb{R}^{\left ( h\times w \right ) \times \left ( h\times w \right )}$ can be seen as the query-key similarities for structure \cite{PAG}, which represent the correlation between image-internal spatial features, influence the structure and shape details of the generated image. In SD, each such spatial feature is indicative of a particular region of the generated image. Inspired by this insight, we achieve object removal by changing the associations between different image-internal spatial features within the self-attention map.

\subsection{Guidance}
A key advantage of diffusion models is the ability to integrate additional information into the iterative inference process for guiding the sampling process, and the guidance can be generalized as any time-dependent energy function from the score-based perspective. Modifying $\epsilon_\theta^{\left ( t \right ) }\left(z_t\right)$ with this energy function can guide the sampling process towards generating samples from a specifically conditioned distribution, formulated as:
\begin{equation}
\hat{\epsilon}_{\theta}^{\left ( t \right ) }\left(z_{t};C\right) =\epsilon_{\theta}^{\left ( t \right ) }\left(z_{t};C\right)-s\;\mathbf{g}\left ( z_t;y \right ), 
\end{equation}
where $C$ represents conditional information, $\mathbf{g}\left ( z_t; y \right ) $ is an energy function and $y$ represents the imaginary labels for the desirable sample and $s$ is the guidance scale. There are many forms of $\mathbf{g}$ \cite{G4,G5,G6,GG1,G2,G3}, the most prevalent of which is classifier-free guidance \cite{G6}, where $C$ represents textual information \cite{liu2023spts,fang2024not,fang2024fewer}, $\mathbf{g}=\epsilon_{\theta}$ and $y=\emptyset$.

\section{Methodology}

\subsection{Overview}
\begin{figure*}[t]
\centering
\includegraphics[width=0.8\textwidth]{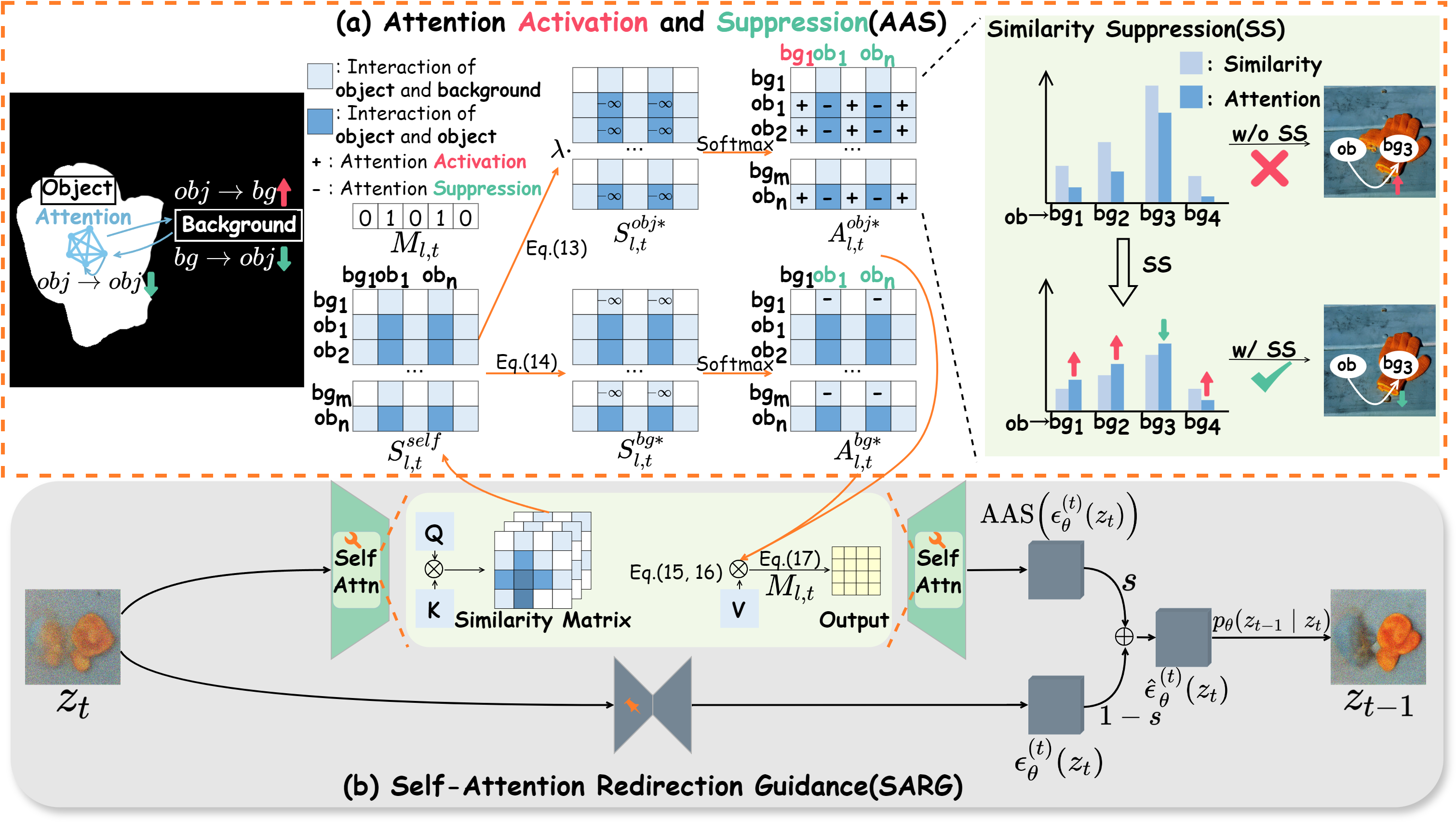} 
\caption{The overview of our proposed Attentive Eraser which consists of two parts: \textbf{(a) Attention Activation and Suppression (AAS)}, a self-attention mechanism modification operation tailored to address the challenges inherent to the object removal task, aims to make the foreground object area's generation more attentive to the background while erasing the object's appearance information. Additionally, Similarity Suppression (SS) serves to suppress the heightened attention to similar objects that may arise due to the inherent nature of self-attention. \textbf{(b) Self-Attention Redirection Guidance (SARG)}, a guidance method applied in the diffusion reverse sampling process, which utilizes redirected self-attention through AAS to guide the sampling process towards the direction of object removal. }
\label{fig2}
\end{figure*}
The overall framework diagram of the proposed method is depicted in Figure~\ref{fig2}. There are two principal components: \textbf{AAS} and \textbf{SARG}, which will be elucidated in more detail in the following sections.

\subsection{Attention Activation and Suppression}

Consider $l$ to be a specific self-attention layer in the U-net that accepts features of dimension $N \times N$, the corresponding similarity matrix and attention map at timestep $t$, $S_{l,t}^{self}, A_{l,t}^{self} \in \mathbb{R}^{N^2 \times N^2}$ can be obtained. The magnitude of the value of $A_{l,t}^{self}\left [ i,j \right ] $ in the self-attention map represents the extent to which the token $i$ generation process is influenced by the token $j$. In other words, row $i$ in the map indicates the extent to which each token in the feature map influences the generation process of token $i$, while column $j$ in the map indicates the extent to which token $j$ influences the generation process of all tokens in the feature map. To facilitate computation and adaptation, we regulate self-attention map $A_{l,t}^{self}$ corporally by changing the similarity matrix $S_{l,t}^{self}$. Specifically, suppose $M_{l,t} \in \mathbb{R}^{ 1 \times N^2}$ is the corresponding flattened mask, among these $N^2$ tokens, we denote the set of tokens belonging to the foreground object region as $F_{l,t}^{obj}$and the set of remaining tokens as $F_{l,t}^{bg}$. Correspondingly, $M_{l,t}$ can be expressed by the following equation:
\begin{equation}
M_{l,t}[i]=\left\{\begin{array}{ll}
1, & i \in F_{l,t}^{obj} \\
0, & i \in F_{l,t}^{bg}.
\end{array}\right.
\end{equation}
We define $S_{l,t}^{obj\rightarrow bg}=\left \{  S_{l,t}\left [ i,j \right ]|i\in F_{l,t}^{obj},j\in F_{l,t}^{bg}\right \}$ to reflect the relevance of the content to be generated in the foreground object area to the background, 
while information about the appearance of the foreground object is reflected in $S_{l,t}^{obj\rightarrow obj}=\left \{  S_{l,t}\left [ i,j \right ]|i\in F_{l,t}^{obj},j\in F_{l,t}^{obj}\right \} $. 
\begin{figure}[t]
\centering
\includegraphics[width=0.9\columnwidth]{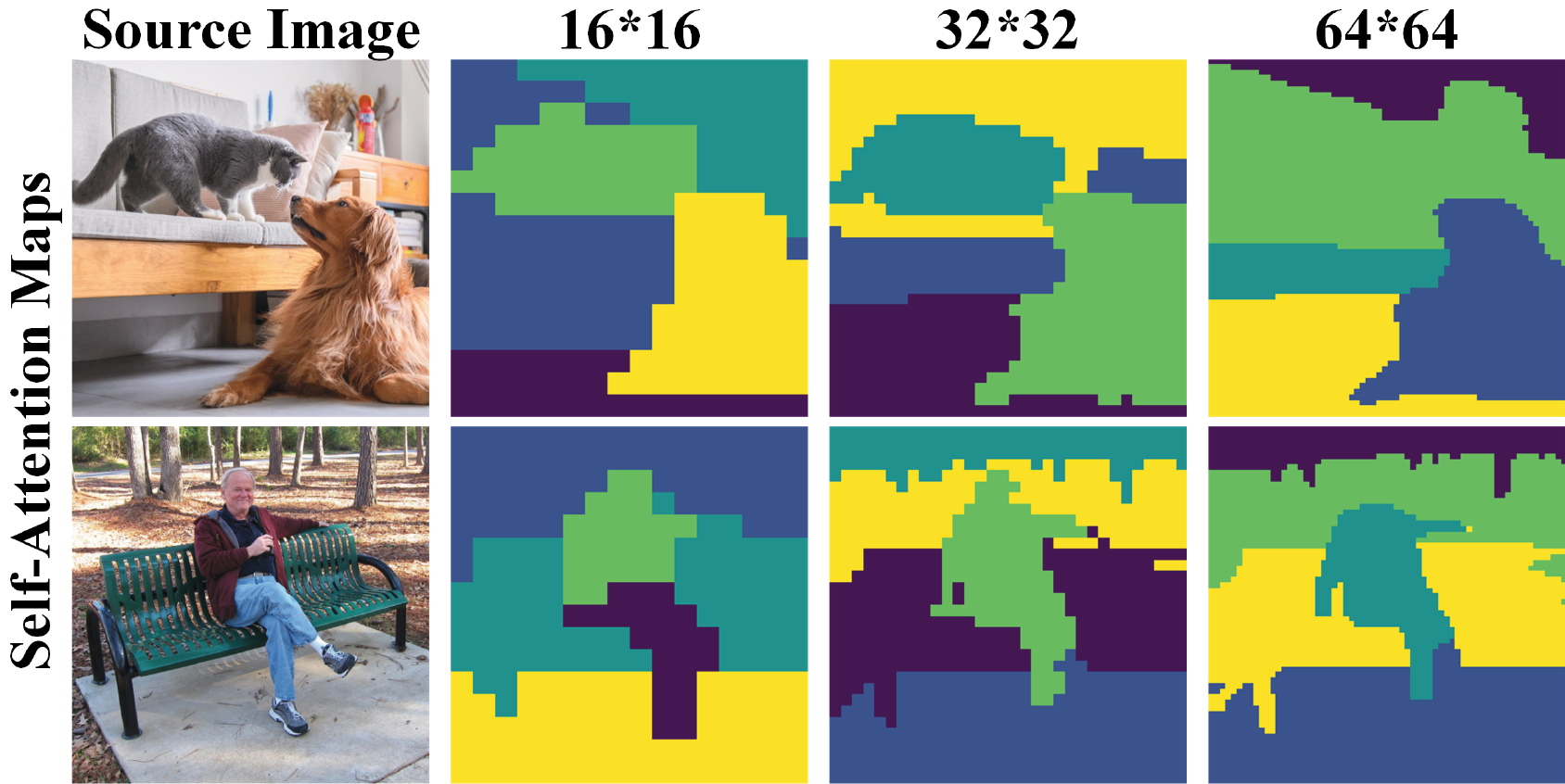} 
\caption{Visualization of the average self-attention maps over all time steps for different layers.}
\label{fig3}
\end{figure}
In the object removal task, we are dealing with foreground objects, and the background should remain the same. As shown in Figure~\ref{fig3}, after DDIM inversion \cite{ddim}, we utilize PCA \cite{pca} and clustering to visualize the average self-attention maps over all time steps for different layers during the reverse denoising process. It can be observed that self-attention maps resemble a semantic layout map of the components of the image \cite{DPL}, and there is a clear distinction between the self-attention corresponding to the generation of the foreground object and background. Consequently, to facilitate object removal during the generation process, an intuitive approach would be to "blend" the self-attention of foreground objects into the background, thus allowing them to be clustered together. In other words, the region corresponding to the foreground object should be generated with a greater degree of reference to the background region than to itself during the generation process. This implies that the attention of the region within the mask to the background region should be increased and to itself should be decreased. Furthermore, the background region is fixed during the generation process and should remain unaffected by the changes in the generated content of the foreground area. Thus, the attention of the background region to the foreground region should also be decreased.

Combining the above analysis, we propose an approach that is both simple and effective: \textbf{AAS} (as shown in Figure~\ref{fig2}(a)). \textbf{Activation} refers to increasing $A_{l,t}^{obj\rightarrow bg}$, which serves to enhance the attention of the foreground-generating region to the background. In contrast, \textbf{Suppression} refers to decreasing $A_{l,t}^{obj\rightarrow obj}$ and $A_{l,t}^{bg\rightarrow obj}$, which entails the suppression of the foreground region's information about its appearance and its effect on the background. Given the intrinsic characteristics of the Softmax function, AAS can be simply achieved by assigning $S_{l,t}^{obj\rightarrow obj}$ to $-\infty$, thereby the original semantic information of the foreground objects is progressively obliterated throughout the denoising process. In practice, the aforementioned operation is achieved by the following equation:
\begin{equation}\label{eq:AAS1}
S_{l,t}^{self\ast }= S_{l,t}^{self} -  M_{l,t}*inf,
\end{equation}
\begin{equation}\label{eq:AAS2}
OP_{l,t}^{\ast }= A_{l,t}^{self\ast}V_{l,t} =\text{softmax}\left ( S_{l,t}^{self\ast}\right) V_{l,t}, 
\end{equation}
where $V_{l,t}$ represents the corresponding value matrix for the time step $t$ of layer $l$.

Nevertheless, one of the limitations of the aforementioned theory is that if the background contains content that is analogous to the foreground object, due to the inherent nature of self-attention, the attention in that particular part of the generative process will be higher than in other regions, while the above theory exacerbates this phenomenon, ultimately leading to incomplete object removal (see an example on the right side of Figure~\ref{fig2}(a)). Accordingly, to reduce the attention devoted to similar objects and disperse it to other regions, we employ a straightforward method of reducing the variance of $S_{l,t}^{obj \rightarrow bg }$, which is referenced in this paper as \textbf{SS}. To avoid interfering with the process of generating the background, we address the foreground and background generation in separate phases:
\begin{equation}
S_{l,t}^{obj\ast}= \lambda S_{l,t}^{self} -M_{l,t}*inf,
\end{equation}
\begin{equation}
S_{l,t}^{bg\ast} = S_{l,t}^{self} -  M_{l,t}*inf,
\end{equation}
\begin{equation}
OP_{l,t}^{obj\ast }= A_{l,t}^{obj\ast}V_{l,t} =\text{softmax}\left ( S_{l,t}^{obj\ast}\right) V_{l,t}, 
\end{equation}
\begin{equation}
OP_{l,t}^{bg\ast }= A_{l,t}^{bg\ast}V_{l,t} =\text{softmax}\left ( S_{l,t}^{bg\ast}\right) V_{l,t},
\end{equation}
where $\lambda$ is the suppression factor less than 1. Finally, to guarantee that the aforementioned operations are executed on the appropriate corresponding foreground and background regions, we integrate the two outputs $OP_{l,t}^{obj\ast }$ and $OP_{l,t}^{bg\ast }$ to obtain the final output $OP_{l,t}^{\ast }$ according to $M_{l,t}^\top$:
\begin{equation}
OP_{l,t}^{\ast }=M_{l,t}^\top\odot OP_{l,t}^{obj\ast }+ \left ( 1-M_{l,t}^\top \right )\odot OP_{l,t}^{bg\ast },
\end{equation}

To ensure minimal impact on the subsequent generation process, we apply SS at the beginning of the denoising process timesteps, for $ t\in \left [ T_I,T_{SS} \right ] $, and still use Eq.\eqref{eq:AAS1}, Eq.\eqref{eq:AAS2} to get output $OP_{l,t}^{\ast }$ for $ t\in \left ( T_{SS},1 \right ] $, where $T_I$ denotes the diffusion steps and $T_{SS}$ signifies the final time-step of SS.
In the following, we denote the U-net processed by the AAS approach as $\text{AAS}\left ( \epsilon_\theta \right ) $.

\subsection{Self-Attention Redirection Guidance}
To further enhance the capability of object removal as well as the overall quality of the generated images, inspired by PAG \cite{PAG}, $\text{AAS}\left ( \epsilon_\theta \right ) $ can be seen as a form of perturbation during the epsilon prediction process, we can use it to steer the sampling process towards the desirable direction. Therefore, the final predicted noise $\hat{\epsilon} _\theta^{\left(t\right)}\left(z_t\right)$ at each time step can be defined as follows:
\begin{equation}
\hat{\epsilon} _\theta^{\left(t\right)}\left(z_t\right)=\epsilon_\theta^{\left(t\right)}\left(z_t\right)+s\left ( \text{AAS}\left ( \epsilon_\theta^{\left(t\right)}\left(z_t\right) \right )-\epsilon_\theta^{\left(t\right)}\left(z_t\right)  \right ), 
\end{equation}
where $s$ is the removal guidance scale. Subsequently, the next time step output latent $z_{t-1}$ is obtained by sampling using the modified noise $\hat{\epsilon} _\theta^{\left(t\right)}\left(z_t\right)$. In this paper, we refer to the aforementioned guidance process as \textbf{SARG}.

Through the iterative inference guidance, the sampling direction of the generative process will be altered, causing the distribution of the noisy latent to shift towards the object removal direction we have specified, thereby enhancing the capability of removal and the quality of the final generated images. For a more detailed analysis refer to Appendix A.

\begin{table*}[t]
\centering
\setlength{\tabcolsep}{4.5pt}       
\resizebox{\textwidth}{!}{
\begin{tabularx}{\textwidth}{c|c|c|c|c|c|c|c|c}
\toprule
Method & Training & Mask & Text & FID$\downarrow$ & LPIPS$\downarrow$ & \textbf{Local FID}$\downarrow$ & \textbf{CLIP consensus}$\downarrow$ & \textbf{CLIP score}$\uparrow$\\
\midrule
SD2.1inp & \usym{2713} & \usym{2713} & \usym{2717} & \textbf{3.805} & 0.3012 & 8.852 & 0.1143 & 21.89\\
SD2.1inp & \usym{2713} & \usym{2713} & \usym{2713} & \underline{4.019} & 0.3083 & 7.194 & 0.1209 & 22.27\\
PowerPaint & \usym{2713} & \usym{2713} & \usym{2717} & 6.027 & \underline{0.2887} & 10.02 & 0.0984 & 22.74\\
Inst-Inpaint & \usym{2713} & \usym{2717} & \usym{2713} & 11.42 & 0.4095 & 43.47 & \underline{0.0913} & 23.02\\
LAMA & \usym{2713} & \usym{2713} & \usym{2717} & 7.533 & \textbf{0.2189} & 6.091 & - & \textbf{23.57}\\
SD2.1+SIP w/o SARG & \usym{2717} & \usym{2713} & \usym{2717} & 5.98 & 0.2998 & 15.58 & 0.1347 & 22.05\\
\textbf{SD2.1+SIP w/ SARG(ours)} & \usym{2717} & \usym{2713} & \usym{2717} & 7.352 & 0.3113 & \underline{5.835} & \textbf{0.0734} & \underline{23.56}\\
\textbf{SD2.1+DIP w/ SARG(ours)} & \usym{2717} & \usym{2713} & \usym{2717} & 7.012 & 0.2995 & \textbf{5.699} & - & 23.43 \\
\bottomrule
\end{tabularx}
}
\caption{Quantitative comparison with other methods. We have indicated in the table whether each method requires training and whether it necessitates mask or prompt text as conditional inputs. In the CLIP consensus metric, deterministic process methods (lacking randomness) are denoted with a '-'. The optimal result and object removal-related metrics are represented in bold, and the sub-optimal result is represented in underlining.}
\label{table1}
\end{table*}

\begin{figure*}[t]
\centering
\includegraphics[width=0.8\textwidth]{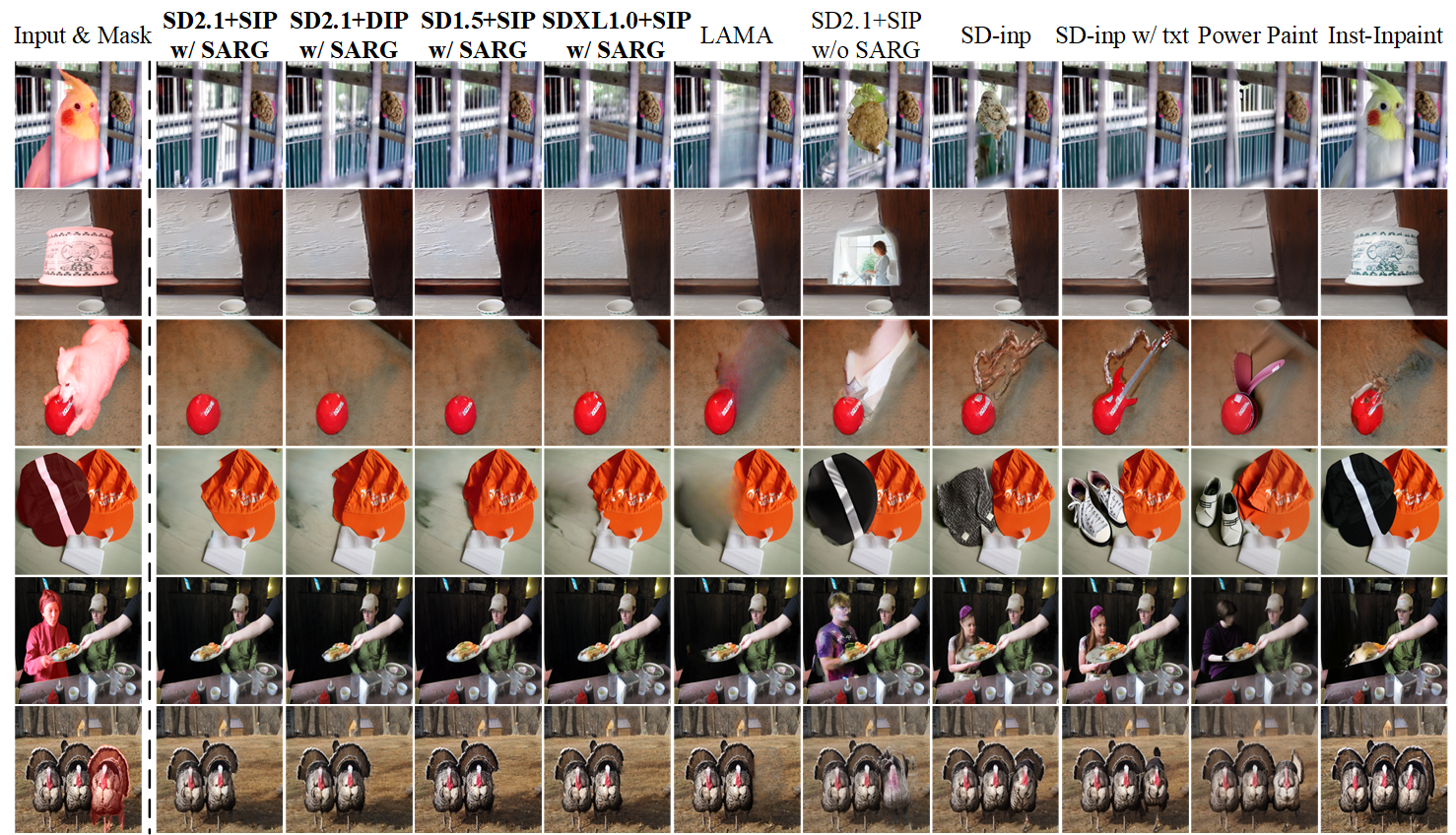} 
\caption{Visual comparison with other methods. The mask is indicated with a red highlight in the input image. Our methods are highlighted in bold.}
\label{Qualitative}
\end{figure*}

\begin{figure}[t]
\centering
\includegraphics[width=0.9\columnwidth]{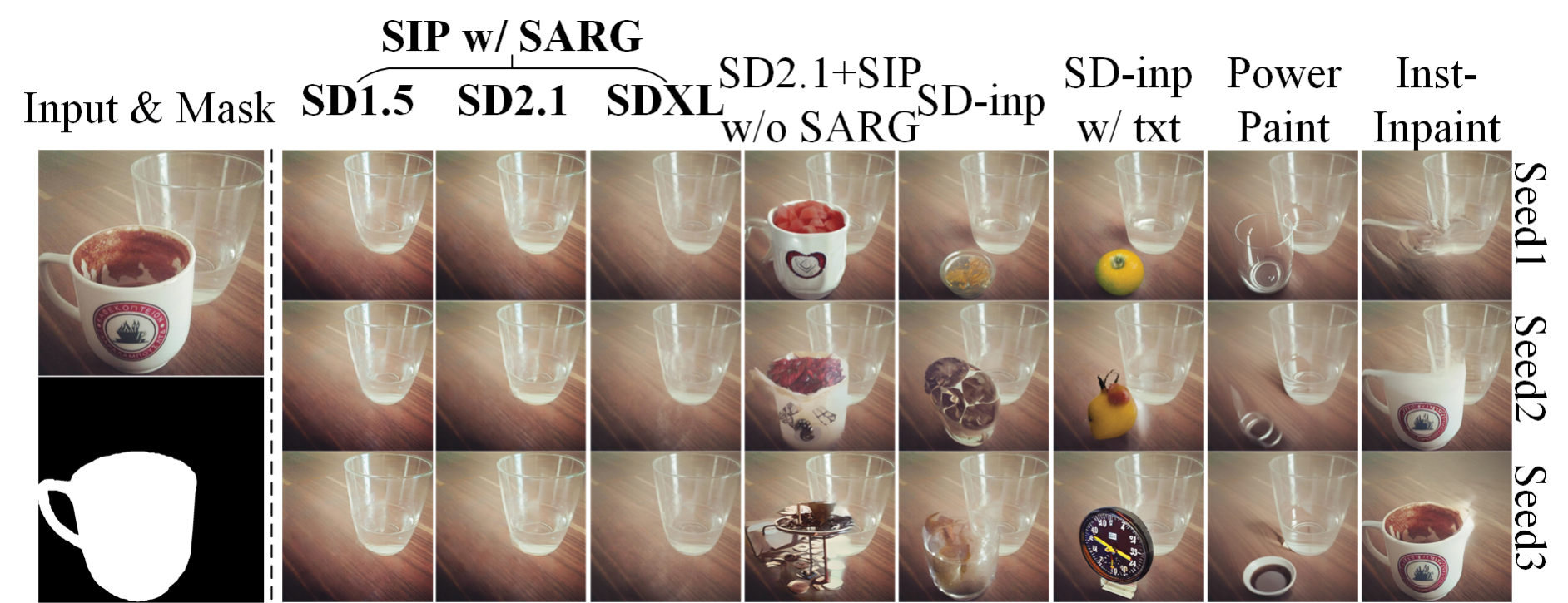} 
\caption{Visual comparison of object removal stability with other methods using three distinct random seeds.}
\label{stable}
\end{figure}

\section{Experiments}
\subsection{Experimental Setup}
\subsubsection{Implementation Details}
We apply our method on all mainstream versions of Stable Diffusion (1.5, 2.1, and XL1.0)  with two prevailing diffusion-based inpainting pipelines \cite{diffedit,bld} to evaluate its generalization across various diffusion model architectures. Based on the randomness, we refer to pipelines as the stochastic inpainting pipeline (SIP) and the deterministic inpainting pipeline (DIP), respectively. Detailed descriptions of SIP and DIP are provided in Appendix B, with further experimental details available in Appendix C.

\subsubsection{Baseline}
We select the state-of-the-art image inpainting methods as our baselines, including two mask-guided approaches SD-Inpaint \cite{sd}, LAMA \cite{lama} and two text-guided approaches Inst-Inpaint \cite{gqa}, Powerpaint \cite{powerpaint}, to demonstrate the efficacy of our method, we have also incorporated SD2.1 with SIP into the baseline for comparative purposes. 

\subsubsection{Testing Datasets}
We evaluate our method on a common segmentation dataset OpenImages V5 \cite{OpenImages}, which contains both the mask information and the text information of the corresponding object of the mask. This facilitates a comprehensive comparison of the entire baseline. We randomly select 10000 sets of data from the OpenImages V5 test set as the testing datasets, a set of data including the original image and the corresponding mask, segmentation bounding box, and segmentation class labels.

\subsubsection{Evaluation Metrics}
We first use two common evaluation metrics \textbf{FID} and \textbf{LPIPS} to assess the quality of the generated images following LAMA\cite{lama} setup, which can indicate the global visual quality of the image. To further assess the quality of the generated content in the mask region, we adopt the metrics \textbf{Local-FID} to assess the local visual quality of the image following \cite{ii2}. To assess the effectiveness of object removal, we select \textbf{CLIP consensus} as the evaluation metric following \cite{paintby}, which enables the evaluation of the consistent diversity of the removal effect. High diversity is often seen as a sign of failed removal, with random objects appearing in the foreground area. Finally, to indicate the degree of object removal, we calculate the \textbf{CLIP score} \cite{CLIP, lu2024improving, liu2024omniclip} by taking the foreground region patch and the prompt "background". The greater the value, the greater the degree of alignment between the removed region and the background, effectively indicating the degree of removal.

\subsection{Qualitative and Quantitative Results}
The quantitative analysis results are shown in Table~\ref{table1}. For global quality metrics FID and LPIPS, our method is at an average level, but these two metrics do not adequately reflect the effectiveness of object removal. Subsequently, we can observe from the local FID that our method has superior performance in the local removal area. Meanwhile, the CLIP consensus indicates the instability of other diffusion-based methods, and the CLIP score demonstrates that our method effectively removes the object and repaints the foreground area that is highly aligned with the background, even reaching a competitive level with LAMA, which is a Fast Fourier Convolution-based inpainting model. Qualitative results are shown in Figure~\ref{Qualitative}, where we can observe the significant differences between our method and others. LAMA, due to its lack of generative capability, successfully removes the object but produces noticeably blurry content. Other diffusion-based methods share a common issue: the instability of removal, which often leads to the generation of random artifacts.
To further substantiate this issue, we conducted experiments on the stability of removal. Figure~\ref{stable} presents the results of removal using three distinct random seeds for each method. It can be observed that our method achieves stable erasure across various SD models, generating more consistent content, whereas other methods have struggled to maintain stable removal of the object.

\subsection{User Study and GPT-4o Evaluation}
Due to the absence of effective metrics for the object removal task, the metrics mentioned above may not be sufficient to demonstrate the superiority of our method. Therefore, to further substantiate the effectiveness of our approach, we conduct a user preference study. Table~\ref{table2} presents the user preferences for various methods, revealing consistent results with the quantitative results and highlighting that our method is strongly preferred over other methods. Furthermore, we design fairly and reasonably prompts,  utilizing GPT-4o \cite{gpt} to conduct a further assessment of object removal performance between our method and the runner-up method LAMA. The results also indicate that our method significantly outperforms LAMA, demonstrating exceptional performance. Please refer to Appendix D for more details and visualizations of user study and GPT evaluation.
\begin{figure}[t]
\centering
\includegraphics[width=0.9\columnwidth]{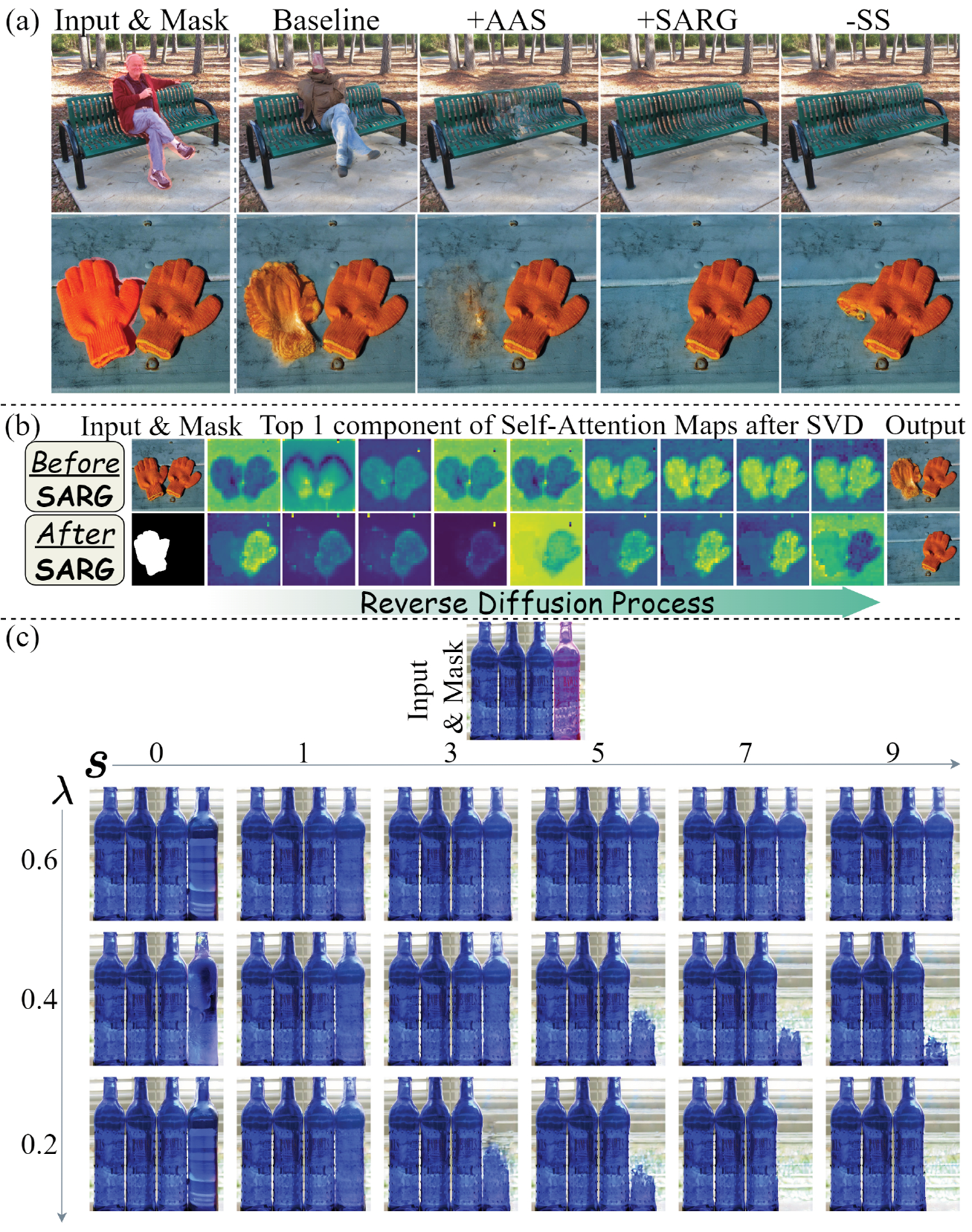} 
\caption{Visualization of ablation experiments on Attentive Eraser.}
\label{ab}
\end{figure}
\subsection{Ablations}

\begin{table}[t]
\centering
\setlength{\tabcolsep}{2pt}

\resizebox{\columnwidth}{!}{
\begin{tabularx}{\columnwidth}{c|c|c}
\toprule
Method & User Study & GPT Evaluation\\
\midrule
SD2.1inp & 10$\%$ & -\\
SD2.1inp(w/ text) & 15.4$\%$ & -\\
PowerPaint & 7.6$\%$ & -\\
Inst-Inpaint & 2.4$\%$ & -\\
LAMA & 19.7$\%$ & 25.53$\%$\\
\textbf{SD2.1+SIP w/ SARG(ours)} & \textbf{44.9$\%$} & \textbf{74.47$\%$}\\
\bottomrule
\end{tabularx}
}
\caption{User study and GPT-4o Evaluation results.}
\label{table2}
\end{table}

To validate the effectiveness of the proposed Attentive Eraser, we conduct ablation studies. We use SD2.1 with SIP as the baseline for comparison,  Figure~\ref{ab} provides a visual representation of the ablation study concerning our method's components. 
Figure~\ref{ab}(a) shows that the application of AAS alone cannot completely remove the foreground object, but integrating it with the sampling process through SARG can effectively remove the object and generate content consistent with the background. At the same time, we also verify the impact of SS, and it can be seen that SS effectively suppresses the generation of similar objects while maintaining the removal efficacy of the general image. As shown in In Figure~\ref{ab}(b), we visualize the heatmaps of the top-1 component of the self-attention maps at each step of the denoising process after SVD \cite{svd}, demonstrating that SARG gradually, as previously stated, "blends" the foreground objects' self-attention into the background to remove objects. In Figure~\ref{ab}(c), we discuss the effect of two parameters (removal guidance $s$ and suppression factor $\lambda$) upon the removal process. It is depicted that as $\lambda$ decreases, the generation of similar objects decreases progressively, thereby reaffirming the efficacy of SS. On the other hand, the intensity of the removal process escalates with an increase in $s$. This suggests that $s$ acts as a pivotal control in modulating the strength of the removal, allowing for a more nuanced and tailored approach to removing objects.

\section{Conclusion}
We present a novel tuning-free method Attentive Eraser, which adeptly harnesses the rich repository of prior knowledge embedded within pre-trained diffusion models for the object removal task. Extensive experiments and user studies demonstrate the stability, effectiveness, and scalability of our proposed method, and also reveal that our method significantly outperforms existing methods.

\section{Acknowledgments}
This work is supported in part by the Summit Advancement Disciplines of Zhejiang Province (Zhejiang Gongshang University - Statistics) and "Digital+" discipline construction management project of Zhejiang Gongshang University (SZJ2022C011).

\clearpage

\bibliography{ref.bib}

\clearpage

\section{Appendix}
For a more thorough comprehension of our method, we have expanded on the details in the ensuing sections.
\subsection{A. Detailed analysis of SARG}
Similar to the derivation in PAG \cite{PAG}, we introduce an implicit discriminator, denoted by $\mathcal{D}$, that distinguishes desirable samples that follow the real data distribution from undesirable samples in the diffusion process. In our work, the original model predictions are regarded as undesirable samples and the AAS-processed model predictions are regarded as desirable samples. The implicit discriminator can be defined as:
\begin{equation}
\mathcal{D}\left ( x_t \right )  =\frac{p\left ( y_{AAS}|x_t \right ) }{p\left ( y|x_t \right )} =\frac{p\left ( y_{AAS} \right ) p\left ( x_t| y_{AAS}\right ) }{p\left ( y\right ) p\left ( x_t| y\right ) } 
\end{equation}
where $y_{AAS}$ and $y$ represent the imaginary labels for the desirable sample and the undesirable sample, respectively.

Subsequently, analogous to WGAN \cite{wgan1,wgan2}, our generator loss for the implicit discriminator, $\mathcal{L}_{\mathcal{G}}$, is established as our energy function and its derivative is calculated as:
\begin{equation}
\begin{aligned}
\nabla_{x_{t}} \mathcal{L}_{\mathcal{G}} & =\nabla_{x_{t}}\left[-\log \mathcal{D}\left(x_{t}\right)\right] \\
& =\nabla_{x_{t}}\left[-\log \frac{p\left ( y_{AAS} \right ) p\left ( x_t| y_{AAS}\right ) }{p\left ( y\right ) p\left ( x_t| y\right ) }\right] \\
& =\nabla_{x_{t}}\left[-\log \frac{p\left(x_{t} | y_{AAS}\right)}{p\left(x_{t} | y\right)}\right] \\
& =-\nabla_{x_{t}}\left(\log p\left(x_{t} \mid y_{AAS}\right)-\log p\left(x_{t} \mid y\right)\right) 
\end{aligned}
\end{equation}

The diffusion sampling process can then be defined as:
\begin{equation}
\begin{aligned}
\hat{\epsilon}_{\theta}\left(x_{t}\right) & =\epsilon_{\theta}\left(x_{t}\right)+s \sigma_{t} \nabla_{x_{t}} \mathcal{L}_{\mathcal{G}} \\
& =\epsilon_{\theta}\left(x_{t}\right)-s \sigma_{t} \nabla_{x_{t}}\left(\log p\left(x_{t} \mid y_{ASS}\right)-\log p\left(x_{t} \mid y\right)\right) \\
& =\epsilon_{\theta}\left(x_{t}\right)+s\left(\text{AAS}( \epsilon_{\theta}\left(x_{t}\right))-\epsilon_{\theta}\left(x_{t}\right)\right)
\end{aligned}
\end{equation}
where the pre-trained score estimation network 
$\epsilon_{\theta}$ and the AAS processed network $\text{AAS}( \epsilon_{\theta})$ are approximations of $- \sigma_{t} \nabla_{x_{t}}\log p\left(x_{t} \mid y\right) $ and $- \sigma_{t} \nabla_{x_{t}}\log p\left(x_{t} \mid y_{ASS}\right) $, respectively.

\subsection{B. Detailed Description of SIP and DIP }
\begin{figure}[t]
\centering
\includegraphics[width=0.9\columnwidth]{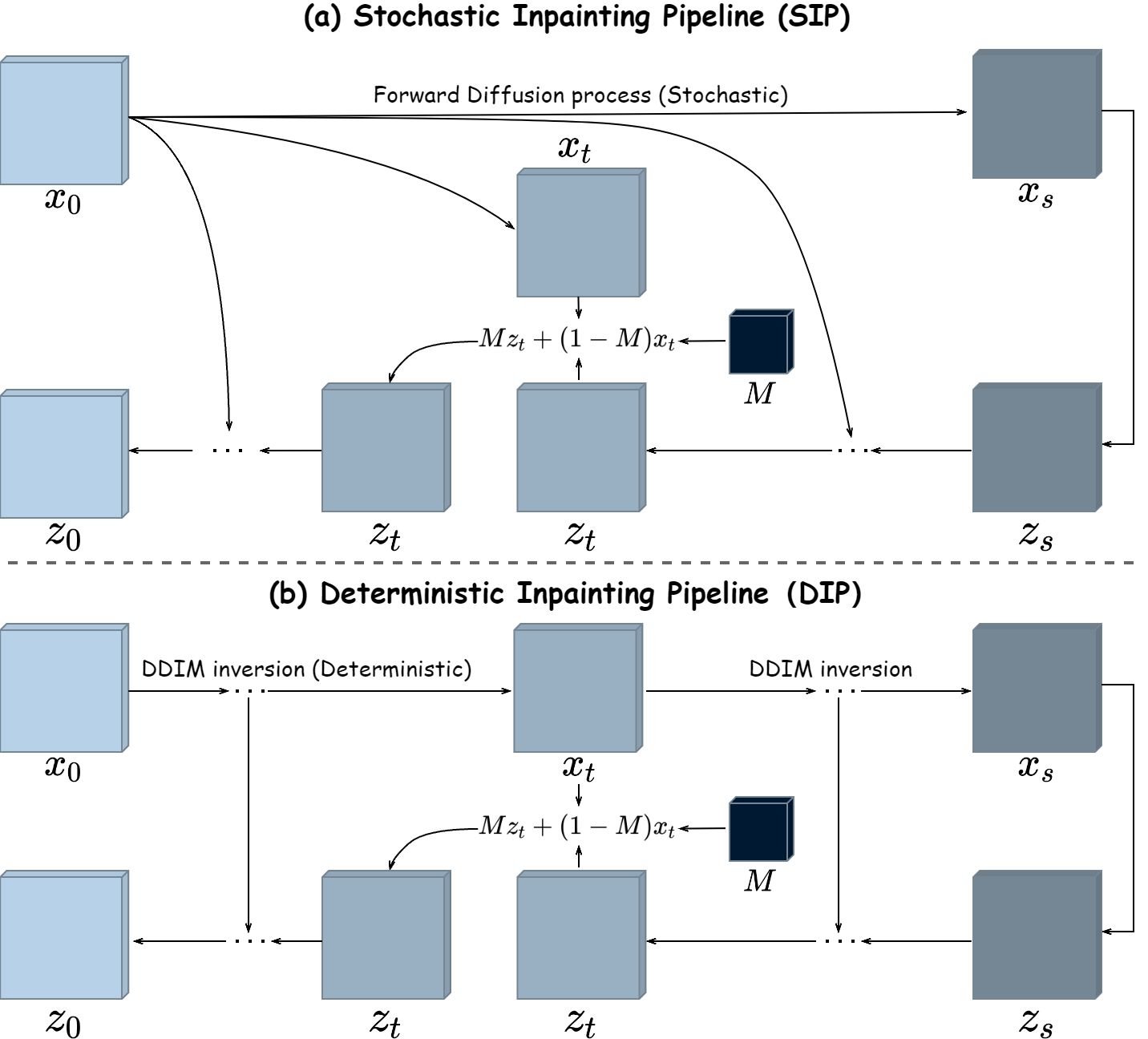} 
\caption{Schematic diagram of SIP and DIP}
\label{sipdip}
\end{figure}

When inpainting real images, two pipelines are commonly employed, which were proposed by BLD \cite{bld} and DiffEdit \cite{diffedit} respectively. We refer to them as the stochastic inpainting process (SIP) and the deterministic inpainting process (DIP) based on the randomness inherent in their processes. The SIP introduces randomness into the generation process by incorporating Gaussian noise. However, the DIP retains the original image information through DDIM inversion \cite{ddim,G5}, which, like DDIM sampling, is a deterministic process and thus does not involve randomness in the generation process. A schematic diagram of SIP and DIP is shown in Figure~\ref{sipdip}. The algorithms of SIP and DIP after applying SARG for object removal are presented in Algorithms ~\ref{alg:algorithm1} and ~\ref{alg:algorithm2}, respectively.

\subsection{C. Additional Experimental Details}
\begin{figure}[t]
\centering
\includegraphics[width=0.9\columnwidth]{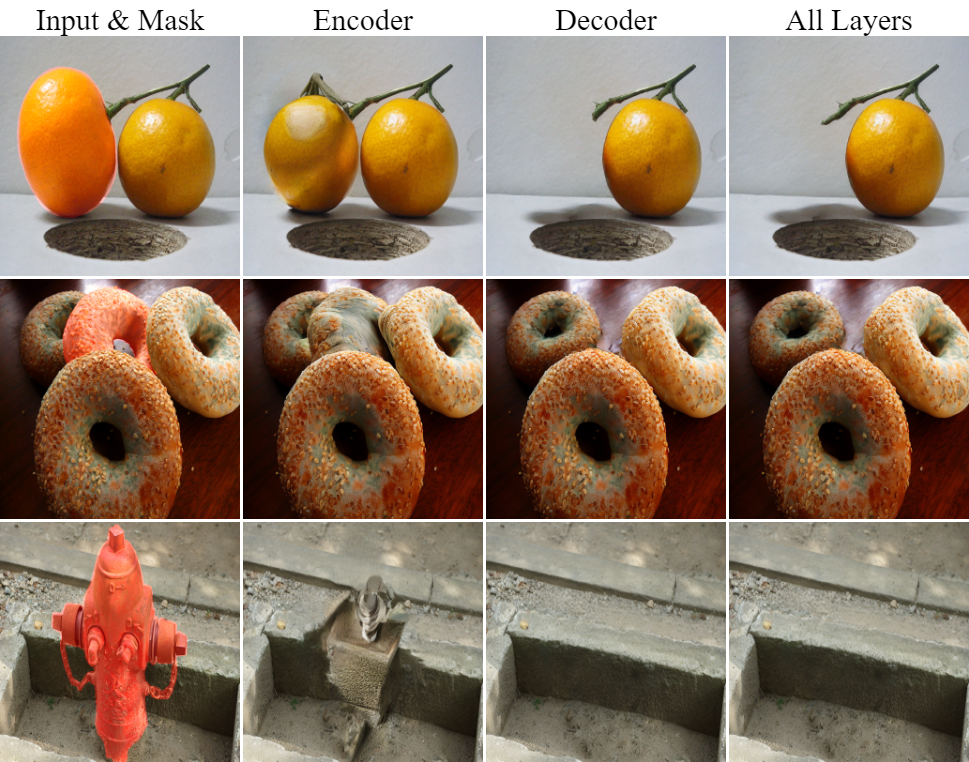} 
\caption{Visual comparative results of applying AAS to different layers of U-net.}
\label{layers}
\end{figure}

Firstly, we present the implementation details of our theory: in all SD models, we adopt DDIM sampling as the default sampling method and \textbf{apply SARG to all time steps}. Concurrently, based on previous research that has established the significant impact of the decoder part in U-net on the appearance information of generated images \cite{de1,de2,de3}, we \textbf{integrate AAS into the decoder of U-net}. We have also provided a visual comparison to substantiate this, as shown in Figure~\ref{layers}. Additionally, starting from the perspective of ensuring the erasure capability and quality as much as possible, we provide the corresponding evaluative parameter settings of SIP and DIP with SARG in Table~\ref{table3}.

\begin{algorithm*}[tb]
\caption{SIP with SARG for Object Removal}
\label{alg:algorithm1}
\textbf{Input}: $X_0$: the input image; $M$: the downsampled input mask; $\epsilon_\theta$: the pre-trained diffusion model; $T_{I}$: the diffusion steps; $\epsilon$: the random Gaussian noise; \\
\textbf{Parameter}: $s$: the removal guidance scale\\
\textbf{Output}: $Z_0$: Edited image after object removal 
\begin{algorithmic}[1] 
\STATE $x_0 \gets \text{VAE-Encoder}\left ( X_0 \right ) $.
\STATE $x_{T_I}\gets \sqrt{\bar\alpha_{T_I} } x_{0}+ \sqrt{\left ( 1-\bar\alpha_{T_I} \right )}\epsilon $.
\STATE $z_{T_I} \gets x_{T_I}$.
\FOR{$t=T_I,...,1$}
\STATE $\hat{\epsilon} _\theta^{\left(t\right)}\left(z_t\right)\gets\epsilon_\theta^{\left(t\right)}\left(z_t\right)+s\left ( \text{AAS}\left ( \epsilon_\theta^{\left(t\right)}\left(z_t\right) \right )-\epsilon_\theta^{\left(t\right)}\left(z_t\right)  \right ) $.
\STATE $z_{t-1} \gets \sqrt{\bar{\alpha}_{t-1} } \left(\frac{z_{t}-\sqrt{1-\bar{\alpha} _{t}} \hat{\epsilon} _\theta^{\left ( t \right ) }\left(z_t\right)}{\sqrt{\bar{\alpha} _{t}}}\right)+\sqrt{1-\bar{\alpha} _{t-1}} \cdot \hat{\epsilon} _\theta^{\left ( t \right ) }\left(z_t\right) $.
\STATE $x_{t-1}\gets \sqrt{\bar{\alpha}_{t-1} }x_0+\sqrt{1-\bar{\alpha}_{t-1}}\epsilon$.
\STATE $z_{t-1}\gets z_{t-1}\odot M+x_{t-1} \odot\left ( 1-M \right )$.
\ENDFOR
\STATE $Z_0 \gets \text{VAE-Decoder}\left ( z_0 \right ) $.
\STATE \textbf{return} $Z_0$
\end{algorithmic}
\end{algorithm*}

\begin{algorithm*}[tb]
\caption{DIP with SARG for Object Removal}
\label{alg:algorithm2}
\textbf{Input}: $X_0$: the input image; $M$: the downsampled input mask; $\epsilon_\theta$: the pre-trained diffusion model; $T_{I}$: the diffusion steps; \\
\textbf{Parameter}: $s$: the removal guidance scale\\
\textbf{Output}: $Z_0$: Edited image after object removal 
\begin{algorithmic}[1] 
\STATE $x_0 \gets \text{VAE-Encoder}\left ( X_0 \right ) $.
\FOR{$t=1,...,T_I$}
\STATE $\left \{ x_t \right \}_{t=1}^{T_I}  \gets \text{DDIM-inv} (x_0)$.
\ENDFOR
\STATE $z_{T_I} \gets x_{T_I}$.
\FOR{$t=T_I,...,1$}
\STATE $\hat{\epsilon} _\theta^{\left(t\right)}\left(z_t\right)\gets\epsilon_\theta^{\left(t\right)}\left(z_t\right)+s\left ( \text{AAS}\left ( \epsilon_\theta^{\left(t\right)}\left(z_t\right) \right )-\epsilon_\theta^{\left(t\right)}\left(z_t\right)  \right ) $.
\STATE $z_{t-1} \gets \sqrt{\bar{\alpha}_{t-1} } \left(\frac{z_{t}-\sqrt{1-\bar{\alpha} _{t}} \hat{\epsilon} _\theta^{\left ( t \right ) }\left(z_t\right)}{\sqrt{\bar{\alpha} _{t}}}\right)+\sqrt{1-\bar{\alpha} _{t-1}} \cdot \hat{\epsilon} _\theta^{\left ( t \right ) }\left(z_t\right) $.
\STATE $x_{t-1}\gets \sqrt{\bar{\alpha}_{t-1} }x_0+\sqrt{1-\bar{\alpha}_{t-1}}\epsilon$.
\STATE $z_{t-1}\gets z_{t-1}\odot M+x_{t-1} \odot\left ( 1-M \right )$.
\ENDFOR
\STATE $Z_0 \gets \text{VAE-Decoder}\left ( z_0 \right ) $.
\STATE \textbf{return} $Z_0$
\end{algorithmic}
\end{algorithm*}

\begin{table}[t]
\centering
\begin{tabular}{c|c|c}
\toprule
Parameters & SIP & DIP\\
\midrule
$T_I$ & 40 & 50\\
$T_{SS}$ & 30 & 40\\
$s$ & 9 & 9\\
$\lambda$ & 0.3 & 0.3\\
\bottomrule
\end{tabular}
\caption{Experiment settings for SIP and DIP with SARG.}
\label{table3}
\end{table}

\begin{figure*}[t]
\centering
\includegraphics[width=0.8\textwidth]{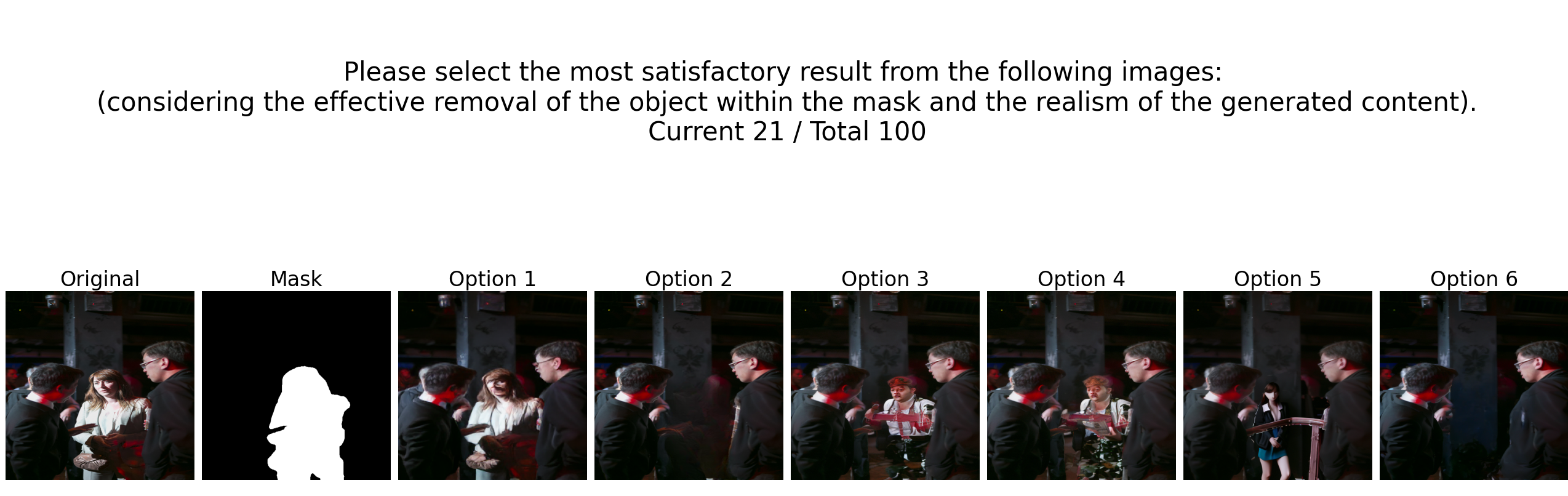} 
\caption{User study print screen.}
\label{userstudy}
\end{figure*}

\begin{figure*}[t]
\centering
\includegraphics[width=0.8\textwidth]{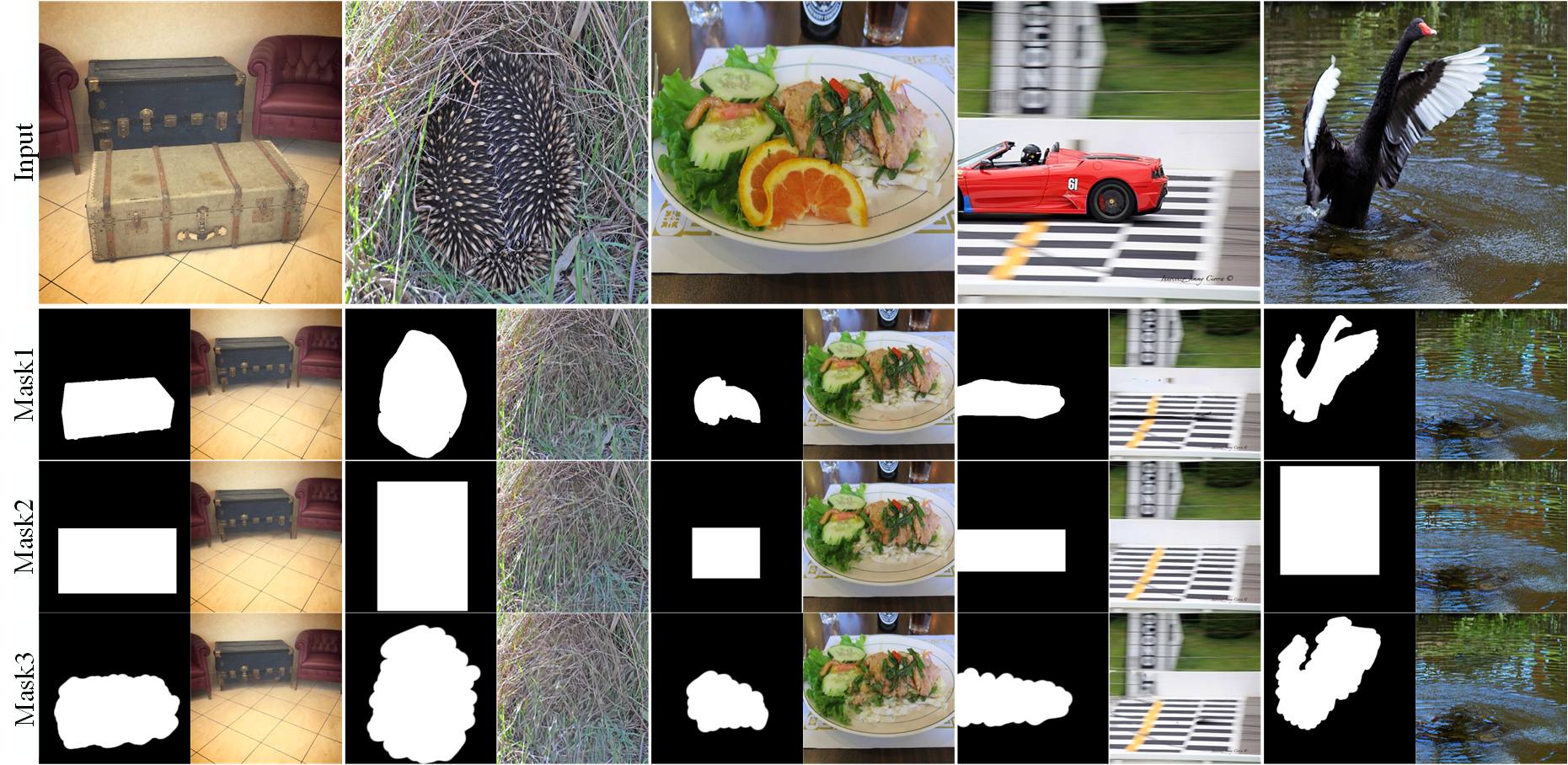} 
\caption{Robustness experiment of our method to input masks on SD2.1 with SIP.}
\label{mt}
\end{figure*}

Below, we will provide a brief overview of the comparative methods in the baseline as well as the corresponding experimental setup:
\begin{itemize}
\item \textbf{SD-Inpaint} is finetuned from Stable Diffusion and is capable of accepting a mask as input for inpainting. In the experiments, we integrate this model with varying input conditions into the baseline, corresponding to scenarios with only mask input and those with both mask and text input (Here "background" is designated as the prompt, while the object label serves as the negative prompt, and the guidance scale is set to 7.5).
\item \textbf{LAMA} leverages Fast Fourier Convolutions (FFCs) to expand the receptive field, along with an effective perceptual loss and aggressive training mask generation strategy, achieving high-quality inpainting on large missing areas. In the experiments, we incorporate the most powerful model as per the official documentation, Big-LAMA, into the baseline.
\item \textbf{Inst-Inpaint} trains a novel conditional diffusion model to implement object removal based on the instructions given as text prompts. In the experiments, we adhere to the original paper settings by designating the text instruction as: "Remove the [object label] at the [location]."
\item \textbf{PowerPaint} achieves versatile and high-quality image inpainting by utilizing learnable task prompts and specialized fine-tuning strategies. In the experiments, we employ the section of the official code related to object removal and following the suggestions provided in their demo, set the guidance scale to 12.
\end{itemize}

During the evaluation process, aside from Inst-Inpaint requiring 256$\times$256 image inputs, we utilize 512$\times$512 images as the input for each method. When calculating the metrics, all output images are resized to 512$\times$512. Given the necessity for the CLIP consensus metric to assess image generation across various random seeds and subsequently calculate the standard deviation of the CLIP embeddings within the foreground object region. We extract 6000 images from the testing dataset of 10000 and generate corresponding results using the random seeds (123, 321, 777) for the computation of this metric. The remaining metrics are calculated using results from the testing dataset of 10000 images generated with the random seed 123.

\begin{figure*}[t]
\centering
\includegraphics[width=0.8\textwidth]{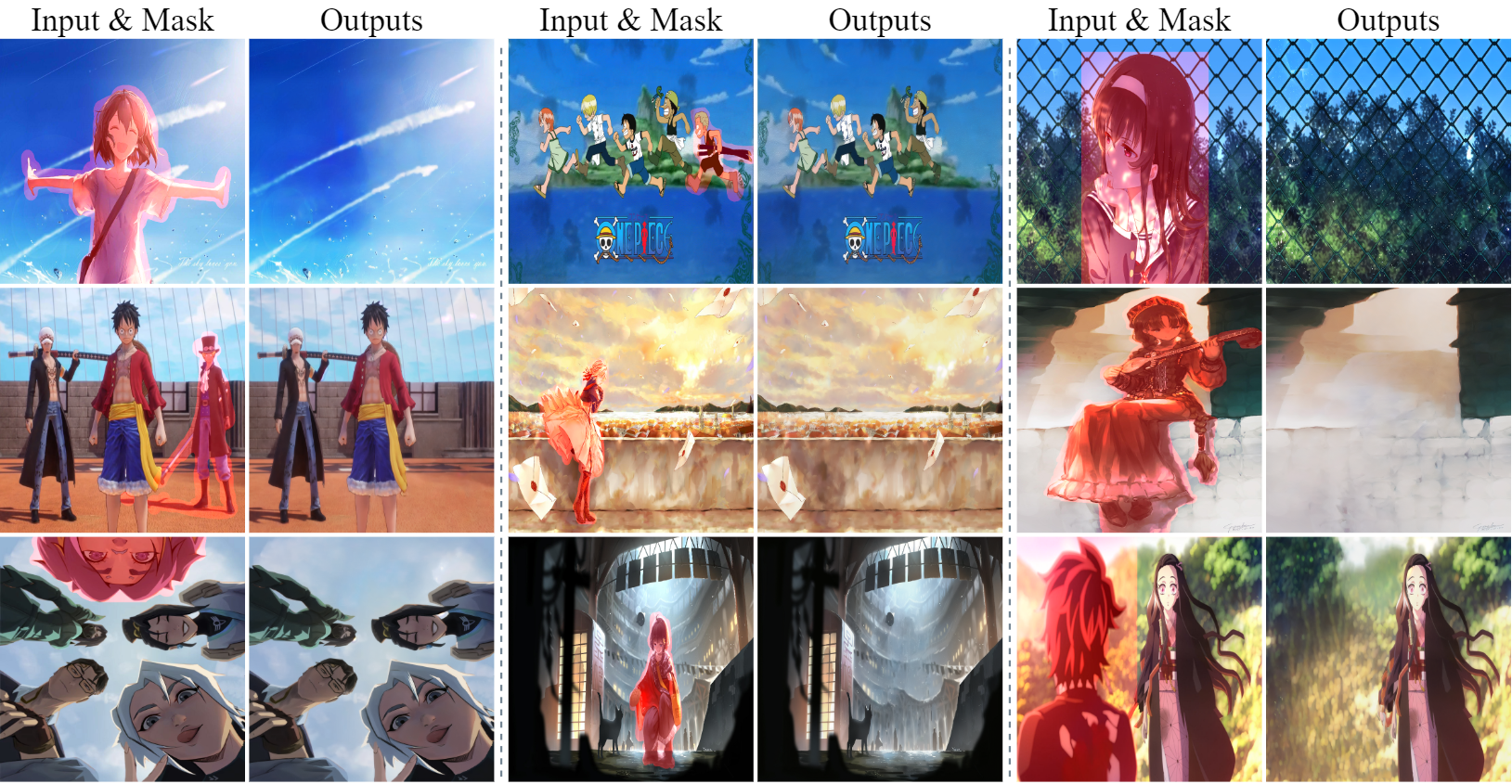} 
\caption{Cartoon image object removal results applying Attentive Eraser on the solarsync model with SIP.}
\label{solarsync}
\end{figure*}

\subsection{D. User Study and GPT-4o Evaluation}
In our user study experiment, we recruited 10 participants to assess each image and determine which one had the best object removal effect based on the provided reference evaluation criteria. Each participant was assigned 100 comparison images obtained through random sampling from the final results. At the same time, we ensured that each round of evaluation was conducted with randomized order and anonymous selection. Finally, we calculated the average user preference percentage for each method. A print screen is provided in Figure~\ref{userstudy}.

In the comparative experiment utilizing GPT-4o against the runner-up LAMA, we tasked GPT with selecting the image with the best object removal effect based on a fairly and reasonably designed prompt. The prompt was as follows: "You are an expert in evaluating generated images. There are two images with their corresponding masks. Please assess the following aspects: 1. Whether the object within the mask has been effectively removed and consistent content with the background has been generated within the mask area. 2. The realism of the generated content within the mask.  Based on these criteria, please tell me which image is better." We conducted experiments with three different random seeds, randomly selecting 1000 pairs of images each time, and finally provided the selection rate based on the results of these 3000 image pairs.

\subsection{E. Robustness and Scalability Analysis}
Furthermore, we demonstrate the robustness of our method to input masks and its scalability to other pre-trained models. As shown in Figure~\ref{mt}, we utilize three mask types varying in refinement levels to assess the robustness of our method: instance segmentation masks, segmentation bounding box masks, and hand-drawn masks. It can be observed that even with the coarse hand-drawn masks, our method effectively removes the target and generates a plausible background, demonstrating that the performance of our method is not hindered by the mask's level of refinement. Additionally, as shown in Figure~\ref{solarsync}, our method is not only applicable to the pre-trained models generating natural images ($i.e.$ SD1.5, 2.1) but can also be extended to models for anime and cartoon images, such as solarsync \cite{solarsync}.

\begin{figure}[t]
\centering
\includegraphics[width=0.9\columnwidth]{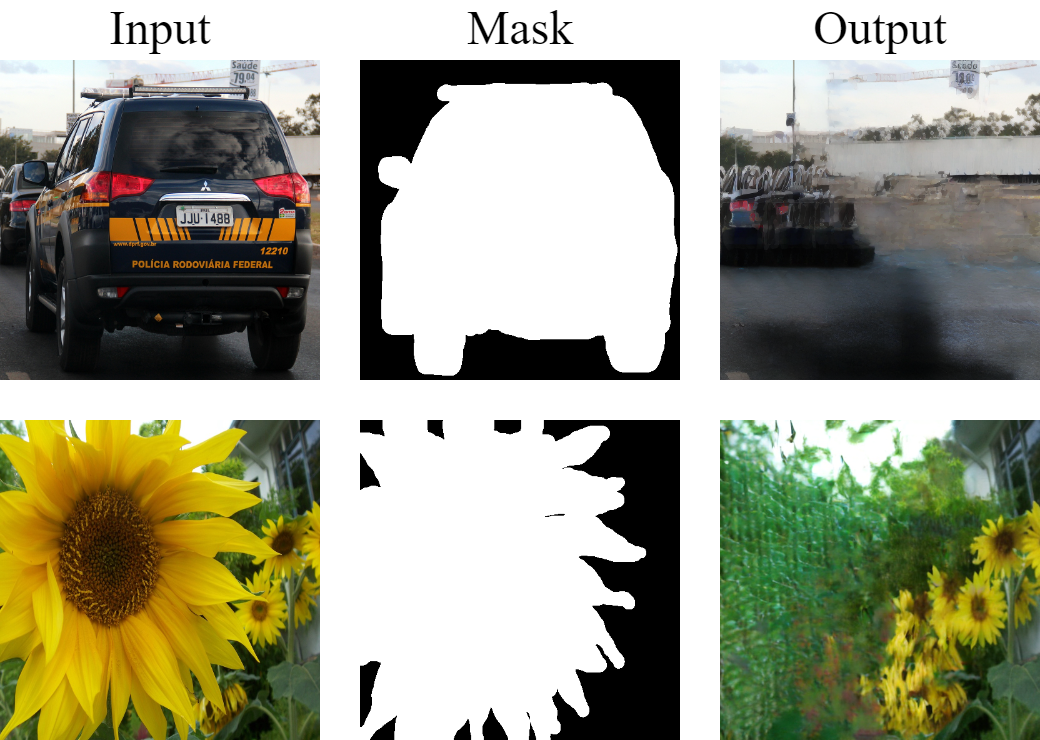} 
\caption{Failure cases.}
\label{fc}
\end{figure}

\subsection{F. Limitations}
Our method has two primary drawbacks. Firstly, it shares a common issue with guidance methods, namely the increased inference time due to the necessity of two times of U-net predicting noise.
Secondly, when the mask area is too large, the scarcity of referable background areas may result in the poor reconstruction of the removal region, leading to the generation of artifacts, as shown in Figure~\ref{fc}. We will endeavour to overcome these limitations in the development of generative AI~\cite{feng2025docpedia,tang2022transcriptioon, tang2023character,tang2022optimal, tang2024textsquare, tang2024mtvqa, zhao2024tabpedia}.

\subsection{G. Additional Results}
\begin{figure*}[t]
\centering
\includegraphics[width=0.8\textwidth]{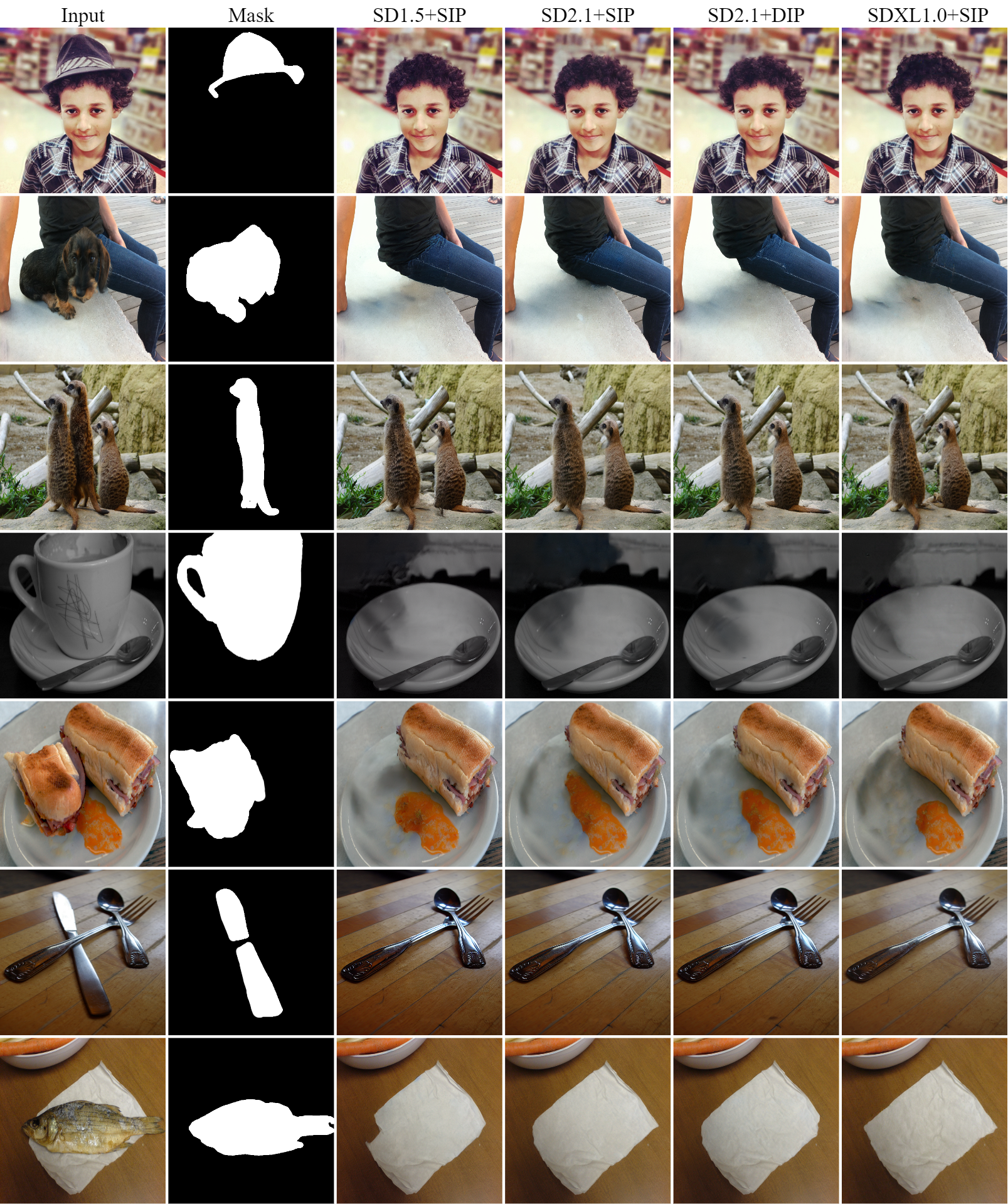} 
\caption{More results of the proposed method.}
\label{ms1}
\end{figure*}

\begin{figure*}[t]
\centering
\includegraphics[width=0.8\textwidth]{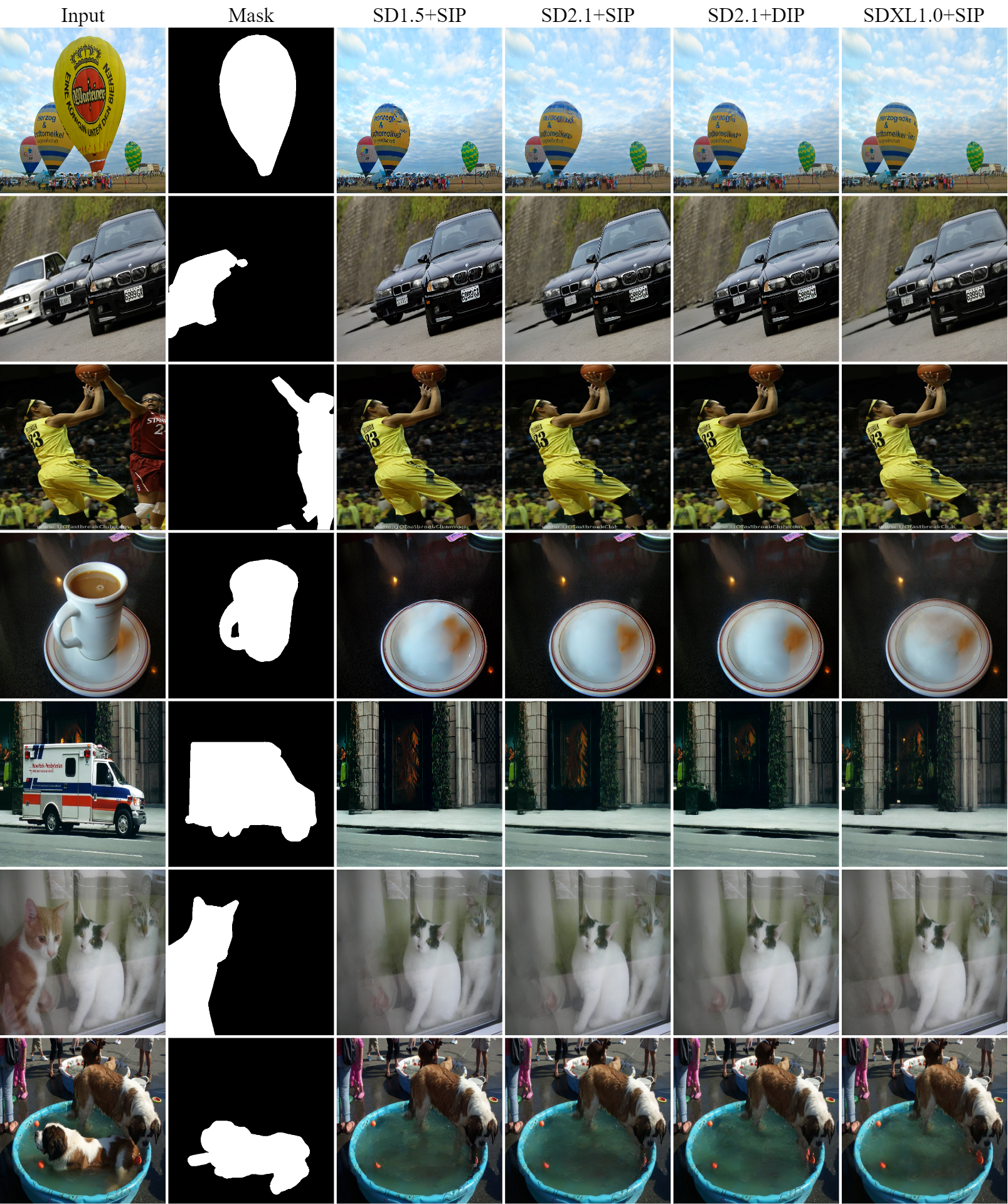} 
\caption{More results of the proposed method.}
\label{ms2}
\end{figure*}

In this section, we provide more samples of the object removal results in Figure~\ref{ms1} and~\ref{ms2}. By applying the Attentive Eraser to various inpainting pipelines and across different SD models, we demonstrate the robustness, effectiveness, and extensibility of our method. It successfully unleashes the potential for object removal in a multitude of pre-trained diffusion models. 

\end{document}